%% file: ref.tex
\documentclass[runningheads]{llncs}
\usepackage{graphicx}

\usepackage{algorithm}
\usepackage{algorithmic}
\usepackage{times}
\usepackage{soul}
\usepackage{url}
\usepackage[utf8]{inputenc}
\usepackage{graphicx}
\usepackage{amsmath}
\usepackage{booktabs}
\usepackage{color}
\usepackage{multirow}
\usepackage{amsfonts}

\usepackage{amssymb}
\usepackage{bbding}
\usepackage{pifont}
\usepackage{wasysym}
\begin{document}
\title{Adaptive Graph Convolution Networks \\ 
for Traffic Flow Forecasting}
\author{Zhengdao Li\inst{1},
Wei Li\inst{2} \and
Kai Hwang\inst{1}
}

\authorrunning{Z. Li et al.}

\institute{The Chinese University of Hong Kong (Shenzhen) \and
 Huazhong University of Science and Technology 
}

\maketitle              % typeset the header of the contribution
\begin{abstract}
Traffic flow forecasting is a highly challenging task due to the dynamic spatial-temporal road conditions. Graph neural networks (GNN) has been widely applied in this task. However, most of these GNNs ignore the effects of time-varying road conditions due to the fixed range of the convolution receptive field. In this paper, we propose a novel Adaptive Graph Convolution Networks (AGC-net) to address this issue in GNN. The AGC-net is constructed by the Adaptive Graph Convolution (AGC) based on a novel context attention mechanism, which consists of a set of graph wavelets with various learnable scales. The AGC transforms the spatial graph representations into time-sensitive features considering the temporal context. Moreover, a shifted graph convolution kernel is designed to enhance the AGC, which attempts to correct the deviations caused by inaccurate topology. Experimental results on two public traffic datasets demonstrate the effectiveness of the AGC-net\footnote{Code is available at: https://github.com/zhengdaoli/AGC-net} which outperforms other baseline models significantly.

\end{abstract}

\input{intro}

\input{relwork}
\input{prel}
\input{method}

\input{exp}
\input{conc}

\bibliographystyle{splncs04}

{\small
\bibliography{ref}}

\end{document}

%% file: intro.tex
\section{Introduction}
Traffic flow forecasting is the task of predicting future traffic flow based on past data \cite{diao2019dynamic}. Accurate and reliable forecasting is crucial for providing drivers and intelligent transportation systems (ITS) with better decision-making abilities and traffic control schemes \cite{wu2016urban}.

Early methods focused on time-series modeling, such as Kalman Filtering \cite{makridakis1997arma}, ARIMA \cite{zivot2006vector}, VAR \cite{chien2003dynamic}, SVR \cite{wu2004travel}, Bayesian models \cite{sun2006bayesian}, etc. These methods mainly rely on statistical models and make independent assumptions about multivariate time-series data from each sensor, making it difficult to capture correlations among variables.

To capture both spatial and temporal information, traffic flow forecasting can be formulated as a multivariate time-series forecasting over a graph constructed from sensor networks. Graph neural networks (GNN) are widely applied in traffic forecasting due to their natural computational architecture for modeling complex spatial relations by aggregating information from neighbors \cite{li2021spatial}. To capture temporal information, most GNN-based models, inspired by recurrent neural networks (RNN), long short-term memory (LSTM), or gated recurrent unit (GRU) \cite{chung2014empirical}, incorporate memory and gated mechanisms \cite{li2017diffusion,bai2020adaptive,zheng2020gman,zhang2021traffic,sun2022ada}.

However, the non-stationary nature of traffic \cite{shekhar2015spatiotemporal,stathopoulos2003multivariate} limits the effectiveness of GNNs due to their fixed receptive fields. This non-stationary property is caused by various factors such as weather conditions, accidents, special events, or road construction, which can cause sudden changes in traffic flow patterns. For instance, accidents on highways can have a more widespread impact than in urban areas, which affect only a localized region. These diverse ranges require graph convolutions to be adaptable to fit these changes. Thus, it is crucial to have models with adaptivity to different traffic scenarios for effective traffic flow prediction. Although some models \cite{Chen2020MRA,guo2021hierarchical} attempt to address this by stacking multiple convolutions with different receptive ranges, they may introduce redundant information due to the lack of a filtering scheme for the receptive fields within a range.

To overcome these challenges, we propose an Adaptive Graph Convolution Networks (AGC-net) that utilizes Adaptive Graph Convolutions (AGC) in each layer. AGC can dynamically filter the best receptive fields locally and globally depending on the traffic flow at each time step. AGC consists of a set of graph wavelets with various learnable scales and a context attention block that refines the best receptive fields according to contextual information. We also introduce a learnable shifted convolution kernel to refine the inaccurate road topology.

Our main contributions lie in two aspects. First, the proposed adaptive graph convolution addresses the limitation of conventional GNN in handling spatial-temporal applications due to the fixed-range receptive fields. Second, extensive experimental results demonstrate that our AGC-net achieves the best performance on both two public datasets.

%% file: relwork.tex
\section{Related work}
\subsection{Traffic Forecasting}
Early traffic forecasting methods are not satisfactory. Because the statistical models they base on, such as Kalman Filtering \cite{makridakis1997arma}, ARIMA \cite{zivot2006vector}, VAR \cite{chien2003dynamic} etc., rely on stationarity assumption and are very limited to handle spatial and temporal road conditions simultaneously. Inspired by convolutional neural networks (CNNs) and recurrent neural networks (RNNs), some deep learning based approaches \cite{deng2016latent,liu2016predicting,zhang2017deep} are proposed to extract more complicated spatial, temporal information and higher nonlinearity. However, these methods are limited by the intrinsic properties of CNNs or RNNs, that are not generically designed for the non-Euclidean data, such as traffic networks, so that capturing spatial information with temporal dependencies is difficult. In other aspect, traffic sample is non-stationary, which means the distribution differs in different time segments \cite{shekhar2015spatiotemporal,stathopoulos2003multivariate}. Consequently, the structural correlations among nodes vary over different distributions. That non-stationary property makes a challenge for those methods that assume a fixed structure through all time steps.

\cite{guo2019attention} proposed ASTGCN which works on highway traffic and uses attention mechanisms on spatial dimension and temporal dimension respectively to capture spatial and temporal dynamics. \cite{Chen2020MRA} proposed MRA-BGCN model, this model incorporates edge correlations in graph structure. \cite{zhang2020spatio} proposed a framework SLC with a generic graph convolutional formulation to capture the spatial information dynamically. Besides the purely GNN-based methods, other recent researches are tend to use fully-connected or transformer architecture to capture both spatial and temporal information. \cite{oreshkin2021fc} utilizes fully connected layers as the graph gate and the time gate to extract spatial and temporal information respectively. In fact, the graph gate also uses an adjacency-like matrix acting as a graph convolution kernel. \cite{xu2020spatial} constructs a spatial transformer and a temporal transformer architectures. In spatial transformer, the learnable positional embedding layer is formulated as an $N\times N$ matrix to learn the spatial correlation.

\subsection{Graph Neural Networks}
There are two mainstream graph neural networks in terms of convolution operator\cite{zhou2020graph}, i.e., spatial-based and spectral-based. Spatial-based GNNs utilize aggregating operation to filter information from node neighborhoods, such an operation is an analogy to a graph convolution, the aggregator is known as a convolution operator  \cite{hamilton2017inductive,niepert2016learning,gao2018large}. Spectral-based GNNs are based on graph spectral theory\cite{hammond2011wavelets}, pioneering works such as \cite{bruna2013spectral} and other following GCNs \cite{bronstein2017geometric,defferrard2016convolutional,kipf2016semi}implement the graph convolution by graph Fourier transform. Xu proposed GWNN \cite{xu2019graph}, another kind of spectral-based GNNs which leverages graph wavelet transform \cite{hammond2011wavelets} as the convolution kernel or convolution operator.

%% file: prel.tex
\section{Preliminaries}

\subsection{Notations and Problem Formulation}

The road network can be represented by an undirected attributed graph $G=(V, E)$, where $V$ is a set of $N=|V|$ nodes and $E$ is a set of edges. The signal at time step $t$ over $G$ is denoted as $\mathbf{X}_t \in \mathbb{R}^{N\times C}$, where $C$ is the number of traffic conditions, e.g., speed, flow, etc. Then the goal is formulated by using past $H$ time steps conditions $\mathcal{X} = (\mathbf{X}_{1}, \mathbf{X}_{2}, \dots, \mathbf{X}_{H}) \in \mathbb{R}^{H \times N \times C}$ to predict next $P$ time steps flows  $\mathcal{Y} = (\mathbf{Y}_{H+1}, \mathbf{Y}_{H+2}, \dots, \mathbf{Y}_{H+P}) \in \mathbb{R}^{P\times N}$.

The adjacency matrix of $G$ is denoted by $\mathbf{A}$. The graph Laplacian matrix of $G$ is defined as $\mathbf{L}=\mathbf{D}-\mathbf{A}$, where $\mathbf{D}$ is a diagonal degree matrix with $D_{ii}=\sum_j A_{ij}$. Then the normalized Laplacian matrix is $\mathbf{L}'=\mathbf{I}_N - \mathbf{D} ^{-1/2} \mathbf{A} \mathbf{D}^{-1/2}$, where $\mathbf{I}_N$ is the identity matrix. The real symmetric matrix $\mathbf{L}'$ has $N$ orthonormal eigenvectors and associated non-negative eigenvalues in diagonal matrix form, denoted as $\mathbf{U}=(\mathbf{u}_1,\mathbf{u}_2,\dots,\mathbf{u}_N)$ and $\Lambda=\mathrm{diag}(\lambda_1,\lambda_2,\dots,\lambda_N)$ respectively, such that $\mathbf{L}'= \mathbf{U}\Lambda \mathbf{U}^{-1}$.

\subsection{Graph Wavelet Transformation} \label{gwt_def}

Graph wavelet transform (GWT) has some benefits compared to graph Fourier transform, such as high efficiency, high sparseness, and localized convolution property \cite{xu2019graph}.
GWT uses a set of graph wavelets as the bases in spectral domain, defined as $\mathbf{\Psi}_s = (\boldsymbol{\psi}_{s,1},\boldsymbol{\psi}_{s,2},\dots,\boldsymbol{\psi}_{s,N})$.
Each wavelet $\boldsymbol{\psi}_{s,i}$ corresponds to a signal on the graph diffused away from node $i$ at scale $s>0$.
Mathematically, $\mathbf{\Psi}_s$ is defined as $\mathbf{\Psi}_s = \mathbf{U}\mathbf{G}_s \mathbf{U}^{-1} $, where $ \mathbf{G}_s = \mathrm{diag}(g(s\lambda_1), \ldots, g(s\lambda_N))$. The $\mathbf{G}_s \in \mathbb{R}^{N \times N}$ is the diagonal scaling matrix, and $g(s\lambda) = e^{s\lambda}$ corresponds to a heat kernel \cite{xu2019graph}.
The graph wavelet transform of signal $\mathbf{f}$ over $G$ is defined as $\mathbf{\Psi}_s^{-1}\mathbf{f}$. Then the convolution between signal $\mathbf{g}$ and $\mathbf{f}$ is defined as: 
\begin{equation}
\begin{aligned}
\mathbf{g}*_{G}\mathbf{f} &= \mathbf{\Psi}_s((\mathbf{\Psi}_s^{-1}\mathbf{g})\odot(\mathbf{\Psi}_s^{-1}\mathbf{f})) = \mathbf{\Psi}_s \mathbf{\Theta} \mathbf{\Psi}_s^{-1}\mathbf{f}
\end{aligned}
\end{equation}

\noindent where $\mathbf{\Theta} \in \mathbb{R}^{N \times N}$ is a learnable diagonal matrix.

%% file: method.tex
\section{Methodology}

\subsection{AGC-net Architecture}
% Given the historical observations
% % $[\mathbf{X}_t], t=[1,\dots,H]$
% $\mathcal{X} = \{\mathbf{X}_t\}_{t=1}^H \in \mathbb{R}^{H \times N \times C}$, our goal is to predict the future traffic flows $\mathcal{Y}  = \{\mathbf{Y}_t\}_{t=H+1}^{H+P} \in \mathbb{R}^{P \times N \times C}$.

\begin{figure}[htbp]
\centering
\includegraphics[width=0.7\textwidth]{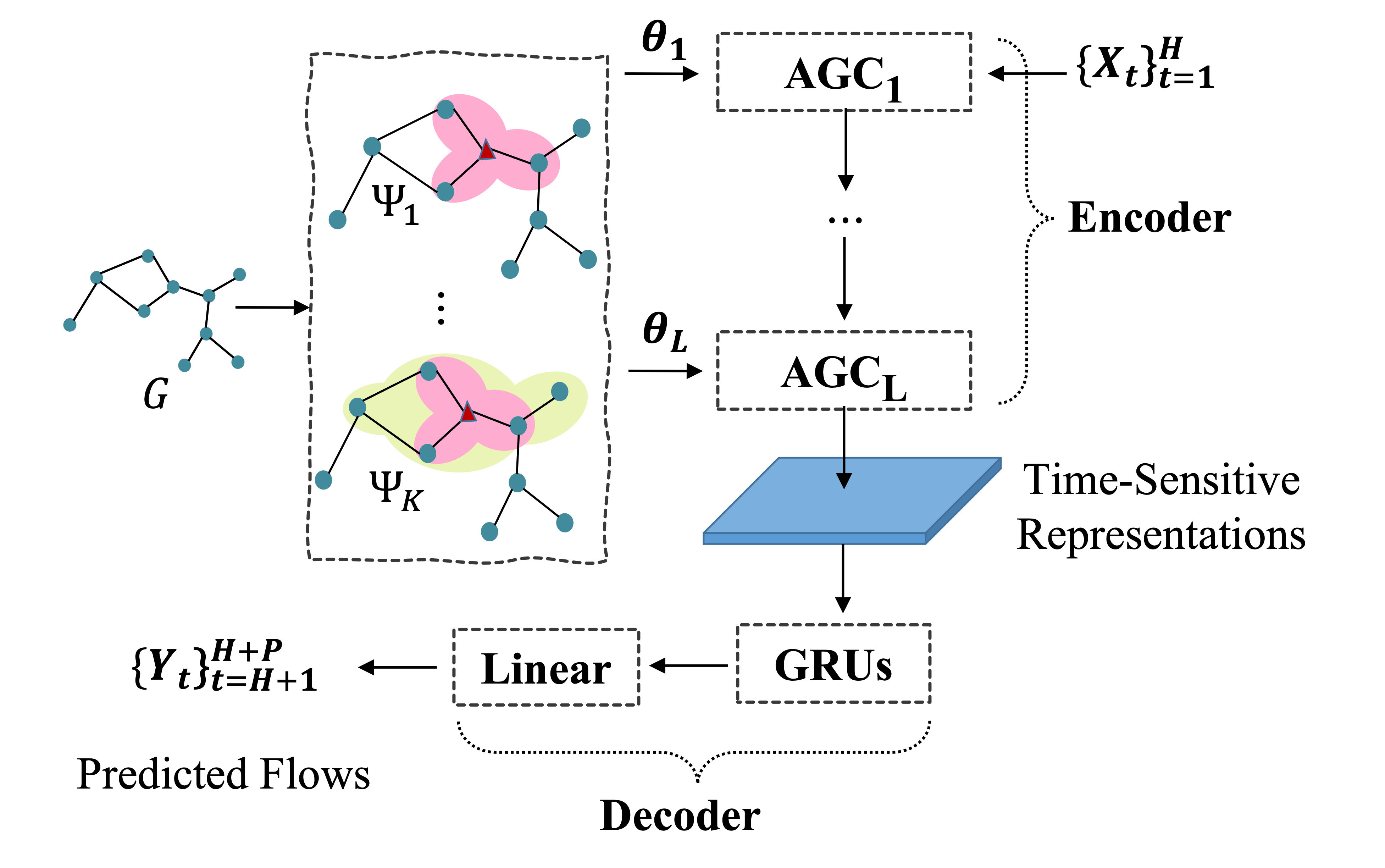}
\caption{\small The architecture of AGC-net. The encoder module is composed of multiple AGC layers. The decoder module leverages GRUs and a linear transformation to produce predicted traffic flows.}
\label{arch}
\end{figure}

As shown in Fig.\ref{arch}, AGC-net is mainly constructed by an encoder model followed by a decoder model.
The goal of encoder is to generate the time-sensitive spatial representations through multiple stacked AGC layers where each AGC takes $K$ graph wavelets $\mathbf{\Psi}_1$ to $\mathbf{\Psi}_K$ and is parameterized independently by $\mathbf{\theta}_1$ to $\mathbf{\theta}_L$ for the construction of convolution kernels. Then the time-sensitive spatial representations are fed into the decoder which is composed of GRU for the prediction.

%%%%%%%%%% figure architecture
%%%%%%%%%% figure architecture

%%%%%%%% AWT
\subsection{Adaptive Graph Convolution Block}
 
To build AGC, the first step is to involve Multi-range Graph Convolution (MGC). The second step is to enhence MGC by a Context Attention Mechanism that can adjust convolutional receptive fields according to the contextual information over time.

\noindent \textbf{Single-range Graph Convolution.} A single-range graph convolution $g_{*k}(\cdot)$ using a graph wavelet $\mathbf{\Psi}_{k}$ defined in Sec.\ref{gwt_def} as the convolution kernel is formulated as: $g_{*k}(\mathbf{X}_t) = \mathbf{\Psi}_{k} \Theta_k \mathbf{\Psi}_{k}^{-1} \mathbf{X}_t\mathbf{W}_k + Bias,$ where $\mathbf{X}_t\in \mathbb{R}^{N \times C_{in}}$ is the input signal at time step $t$, $\mathbf{W}_k \in \mathbb{R}^{C_{in}\times C_{out}}$ is the learnable parameter matrix for feature transformation, $C_{in}$ is the input feature dimension, and $C_{out}$ is the output feature dimension. $\Theta_k \in \mathbb{R}^{N \times N}$ is a learnable diagonal matrix and $Bias \in \mathbb{R}^{N \times C_{out}}$ is the bias matrix.

\noindent \textbf{Multi-range Graph Convolution.} MGC consists of multiple single-range graph convolutions with different kernels. We define an operation $\mathbf{MGC}[g_{*1},g_{*2},...]$ to compose $K$ different graph convolutions. In general, $\mathbf{MGC}$ could be a concatenate operation, i.e., $||_{k=1}^{K}g_{*k}(\cdot)$, or summation of all $g_{*k}(\cdot)$, i.e., $\sum_{k=1}^{K}g_{*k}(\cdot)$. Intuitively and empirically, we find that a learnable coefficient ${\pi}_{k}$ for each $g_{*k}(\cdot)$ can improve the performance, i.e.,

\begin{equation}
\mathbf{MGC}[g_{*1}, \ldots, g_{*K}]= \sum_{k=1}^K \mathbf{\pi}_k \cdot g_{*k}(\cdot).
\end{equation}

Each kernel of $g_{*k}$ corresponds to a $\mathbf{\Psi}_{k} \in \{\mathbf{\Psi}_{k}\}_{k=1}^K$. $K$ indicates the capacity to capture various spatial correlations with different ranges. Naturally, our concern becomes how to find the best ${\pi}_{k}$.

\begin{figure}[t]
\centering
\includegraphics[width=0.7\textwidth]{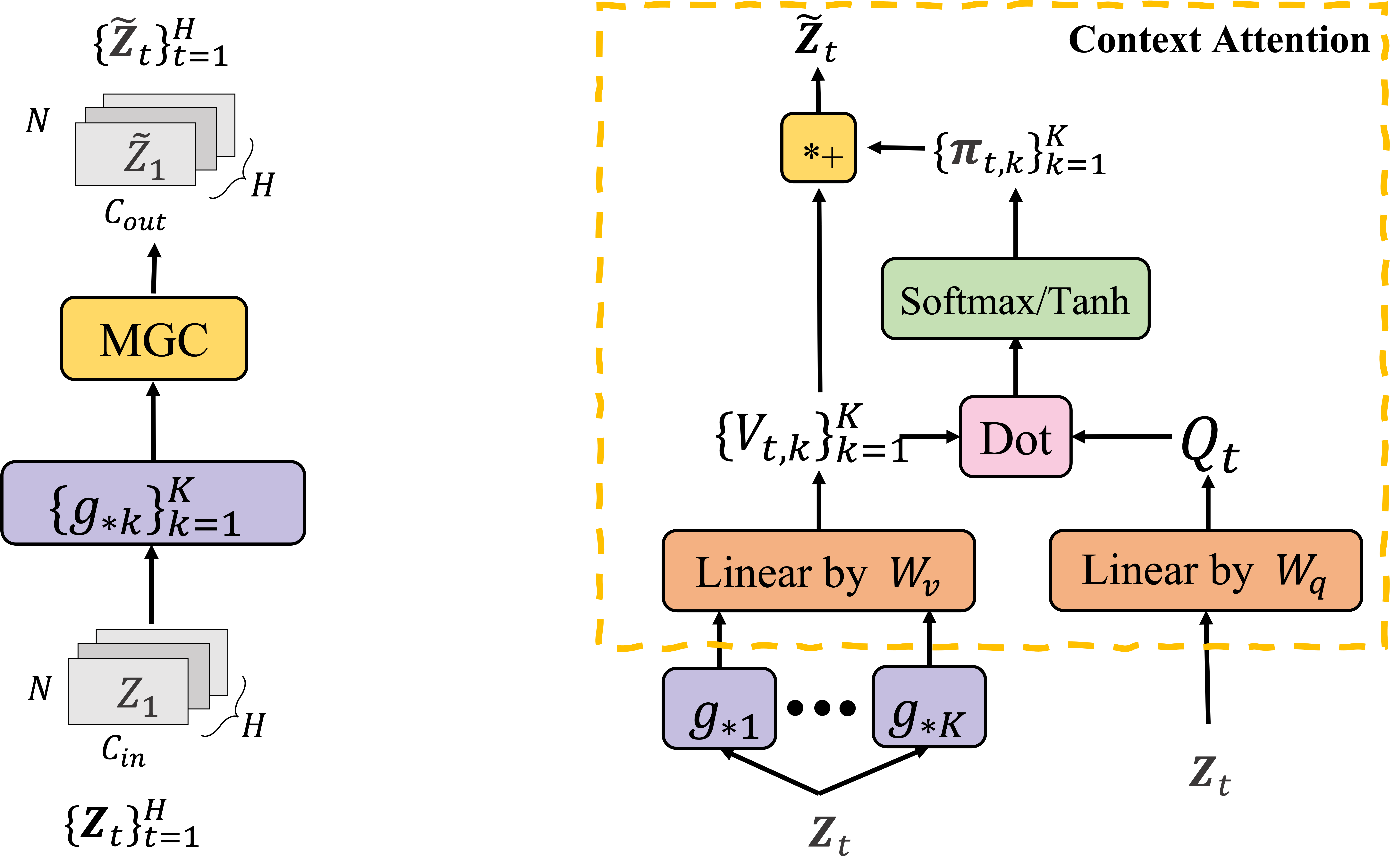}
\caption{\small Multi-range graph convolution with the context attention mechanism.}
\label{multi_att}
\end{figure}

\noindent \textbf{Context Attention Mechanism.} To further leverage the contextual information, we propose the context attention mechanism to learn  $\pi_k$. As shown in Fig.\ref{multi_att}, the MGC with this mechanism (MGC-Attention) at the $l$-th layer takes the last layer hidden state $\mathbf{Z}_t^{l-1}$ and the output of the $g_{*k}(\mathbf{Z}_t^{l-1})$ as the contextual information. Then these two inputs are fed into two independent Linear transformation modules and are transformed into two corresponding representations both in dimension $S$, i.e., $\mathbf{V}_{t,k}=\mathbf{Linear}_{W_v}\left(g_{*k}(\mathbf{Z}_t^{l-1}) \right)$, and $\mathbf{Q}_t=\mathbf{Linear}_{W_q}\left(Z_t^{l-1} \right)$. The normalized inter product divided by dimension scale $S$ of these two representations are calculated as the similarity scores between graph convolutions and the contextual information, i.e.,
 
\begin{equation}
\mathbf{s}_{t, k}=\frac{\mathbf{Q}_t^T \mathbf{V}_{t,k}}{S |\mathbf{Q}_t||\mathbf{V}_{t,k}|}.
\end{equation}

Then we use a \emph{softmax} function to calculate the $\pi_{t, k}$:
\begin{equation}
\pi_{t, k} = \frac{\exp(\mathbf{s}_{t,k})}{\sum_{m=1}^K \exp(\mathbf{s}_{t,m})}.
\end{equation}

%%%%%%%%%%%%

All the computations of $\mathbf{s}_{t, k}$ and $\pi_{t, k}$ can be implemented efficiently by matrix multiplication.

\noindent \textbf{Adaptive Graph Convolution Layer.} We call the $l$-th layer of AGC-net as an adaptive graph convolution layer (AGC), which consists of an MGC-Attention and is simply followed by a non-linear activation function such as $\mathbf{ReLU}(\cdot)$, i.e., $
    \mathbf{AGC}: \mathbf{Z}_t^l \leftarrow \mathbf{ReLU} \left(\sum_{k=1}^{K} \pi_{t, k} g_{*k}(\mathbf{Z}_t^{l-1}) \right). \label{func_AWT}$

Moreover, in our experiments, we compare the MGC-Attention with the MGC-Weighted which uses learnable scalars as the $\pi_{t, k}$ without any contextual information.

%%%%%%%% AWT
%%%%%%%%%%%%%%%%%%    attention illustrationt

\subsection{Learnable Shifted Convolution Kernel}
We design a learnable matrix $\mathbf{D} \in \mathbb{R}^{N \times N}$ for learning a better topology, which can make up for the inaccurate topology in practice that hinders the graph convolution. To address the computational concerns, we use a low rank matrix instead, that is factorized by the product of two low dimension matrices, i.e., $\widetilde{\mathbf{D}}=L1L2$, where $L1 \in \mathbb{R}^{N\times r}, L2 \in \mathbb{R}^{r\times N}, N >> r$. The enhanced graph convolution $g_{*k}$ is simply obtained by replacing the $\mathbf{\Psi}_{k} \Theta_k \mathbf{\Psi}_{k}^{-1}$ with $\mathbf{\Psi}_{k} \Theta_k \mathbf{\Psi}_{k}^{-1}+\alpha  \widetilde{\mathbf{D}},$ where the $\alpha$ is a hyper-parameter to adjust the contribution of $\widetilde{\mathbf{D}}$. The topology is very sparse in practice. To guarantee the sparsity, the Frobenius norm $||\widetilde{\mathbf{D}}||_F^2$ is introduced in the loss function.

%% file: exp.tex
%%%%%%%% performance table
\begin{table*}[t]
\centering
\small
\caption{Performance comparison of our method with seven others.}
\begin{tabular}{c|c|ccc|ccc|ccc}
\toprule[1pt]
\multicolumn{1}{c|}{} & \multicolumn{1}{c|}{} & \multicolumn{3}{c|}{Prediction of 15 min} & \multicolumn{3}{c|}{Prediction of 30 min } & \multicolumn{3}{c}{Prediction of 1 hour} \\
% \cline{3-11}
\cmidrule{3-11}
\multicolumn{1}{c|}{\multirow{-2}{*}{Dataset}} & \multicolumn{1}{c|}{\multirow{-2}{*}{Models}} 
& MAE & RMSE & MAPE &MAE & RMSE & MAPE &MAE & RMSE & MAPE \\

\midrule
\multirow{6}*{METR-LA}
&ARIMA &3.99 & 8.12 & 9.6\% &5.15 & 10.45 & 12.7\% &6.90 & 13.23 & 17.4\% \\
&DCRNN&2.77 & 5.38 & 7.3\% &3.15 & 6.45 & 8.8\% &3.60 & 7.59 & 10.5\% \\
&Graph WaveNet&2.69 & 5.15 & 6.9\% &3.07 & 6.22 & 8.4\% &3.53 & 7.37 & 10.0\% \\
&SLCNN& \textbf{2.53} & 5.18 & 6.7\% &\textbf{2.88} & 6.15 & \textbf{8.0}\% &\textbf{3.30} & 7.20 & 9.7\% \\
&MRA-BGCN& 2.67 & 5.12 & 6.8\% & 3.06 & 6.17 & 8.3\% & 3.49 &7.30 & 10.0\% \\
&STAWnet&2.70 & 5.22 & 7.0\% & 3.04 & 6.14 & 8.2\% &3.44 & 7.16 & 9.8\% \\
&FC-GAGA& 2.70 & 5.24 & 7.0\% & 3.04 & 6.19 & 8.3\% & 3.45 &7.19 & 9.9\% \\
&\textbf{AGC-net(Ours)}& 2.61 & \textbf{4.83} & \textbf{6.6\%} & 2.94 & \textbf{5.68} & \textbf{8.0\%} & 3.34 & \textbf{6.61} & \textbf{9.4\%} \\
& \textbf{Improvements(\%)}&  - &  \textbf{+4.7} & \textbf{+2.9} &  - & \textbf{+7.5}&  - &  - & \textbf{+7.7}& \textbf{+3.1} \\

\midrule

\multirow{6}*{PeMS-BAY}
&ARIMA&1.62 & 3.30 & 3.5\% &2.33 & 4.76 & 5.4\% &3.38 & 6.50 & 8.3\% \\
&DCRNN&1.38 & 2.95 & 2.9\% &1.74 & 3.97 & 3.9\% &2.07 & 4.74 & 4.9\% \\
&Graph WaveNet&1.30 & 2.74 & 2.7\% &1.63 & 3.70 & 3.7\% &1.95 & 4.52 & 4.6\% \\
&SLCNN& 1.44 & 2.90 & 3.0\% &1.72 & 3.81 & 3.9\% &2.03 & 4.53 & 4.8\% \\
&MRA-BGCN&1.29 & 2.72 & 2.9\% & 1.61 & 3.67 & 3.8\% &1.91 &4.46 &4.6\% \\
&STAWnet&1.31 & 2.78 & 2.8\% & 1.62 & 3.70 & 3.7\% &1.89 & 4.36 &4.5\% \\
&FC-GAGA& 1.34 & 2.82 & 2.8 & 1.66 & 3.75 & 3.7\% & 1.93 &4.40 & 4.5\% \\
&\textbf{AGC-net(Ours)}& \textbf{1.18} & \textbf{2.31} & \textbf{2.3\%} & \textbf{1.48} & \textbf{3.14} & \textbf{3.1\%} &\textbf{1.85} & \textbf{3.9} & \textbf{4.2\%} \\
& \textbf{Improvements(\%)}&\textbf{+8.5}& \textbf{+15.1}&\textbf{+20.7}&\textbf{+8.1}&\textbf{+14.4}&\textbf{+18.4}&\textbf{+2.1}&\textbf{+10.6}& \textbf{+8.7}  \\

\bottomrule[1pt]
\end{tabular}
\centering
\label{main_result}
\end{table*}
%%%%%%%% performance table end ------

\section{Experiments}
\subsection{Datasets and Performance Metric}

We conduct experiments on two public traffic datasets. One is \textbf{METR-LA}, which is collected from observation sensors in the highway of Los Angeles County. This dataset uses 207 sensors and 4 months of data dated from 1st Mar 2012 until 30th Jun 2012. The other is \textbf{PeMS-BAY}, which is collected from Caltrans PeMS in Bay Area of California. PeMS-BAY has 6 months of data from 325 sensors (nodes), which ranges from Jan 2017 to May 2017. For both datasets, we adopt three widely used metrics to evaluate the performance of our model, i.e., Mean Absolute Error (\textbf{MAE}), Root Mean Squared Error (\textbf{RMSE}), Mean Absolute Percentage Error (\textbf{MAPE}).

\subsection{Baseline Models}

We compare our AGC-net with the following baseline methods that use the same datasets as ours, i.e., \textbf{ARIMA} \cite{zivot2006vector}. \textbf{DCRNN} \cite{li2017diffusion} combines recurrent neural networks and diffusion convolution. \textbf{Graph WaveNet} \cite{Wu2019GWN} develops an adaptive dependency matrix to capture spatial dependency and stack dilated 1D convolution to capture temporal dependency. \textbf{SLCNN} \cite{zhang2020spatio} uses two structure learning convolutions (using a fixed structure and a learnable structure respectively) to capture the global and local information. \textbf{MRA-BGCN} \cite{Chen2020MRA} incorporates graph edges information in addition to node information to capture spatial dependency. \textbf{STAWnet} \cite{tian2021spatial} utilizes the attention mechanism to directly learn an adjacency matrix for the graph convolution and use temporal convolution networks to capture the temporal information. \textbf{FC-GAGA} \cite{oreshkin2021fc} uses a learnable fully connected hard graph gating mechanism to learn the spatial-temporal information without any prior knowledge of topology.

\subsection{Training Setup}\label{train_setup}
The loss function is MAE with the Frobenius norm of $\widetilde D$. To analyze the effects of some key settings, e.g., wavelet amount, low rank matrix dimension, etc., we conduct several ablation studies which show that for different datasets, the best hyperparameters are different. For example, the best number of graph wavelets for METR-LA dataset is 20, but for PeMS-BAY, it is 15. For the shifted matrices, we set $r$=30 for both METR-LA and PeMS-BAY with $\alpha=0.01$.
We conduct all the experiments on one TeslaV100, using the Adam optimizer \cite{kingma2014adam} with the learning rate $0.002$, weight\_decay $0.0001$, and batch size $128$ to train the model. Further discussions about the effects of these hyperparameters are discussed in the section \ref{effective}. We also try the procedure in \cite{guo2019attention} to involve parallel periodic traffic patterns, i.e., hourly, daily-periodic and weekly-periodic flows.

%  & $A$ & Wavelet & Fusion & Weighted-MGC & Attention-MGC & Shifted Kernel
% ---------------- ablation end -----

\subsection{Performance Analysis}

As shown in Table \ref{main_result}, our method outperforms the other methods in terms of all the metrics on PeMS-BAY and more than half of metrics on METR-LA. In particular, our model improves significantly on RMSE and MAPE metrics, ranging from 10.6\% to 15.1\%, 8.7\% to 20.7\% respectively.

Another observation is that the improvements of our method are different over three metrics and different datasets. This is due to the intrinsic distribution of data. PeMS-BAY has more nodes and edges than METR-LA, which leads to the requirement of extracting more complex spatial information.

% \begin{figure*}[h]
%     \centering
%     \begin{subfigure}{0.4\textwidth}
%     \begin{tabular}{c|c|ccc}
%     \toprule[1pt]
%     Setting \# & Modules &MAE & RMSE & MAPE  \\
%     \hline
%     (a)& \ding{182} &3.70 & 7.14 & 10.7\%  \\
%     (b)&\ding{182}\ding{183}  & 3.53 &  6.87 &  10.3\% \\
%     (c)&\ding{183}\ding{184} &3.58 & 6.89 & 10.5\% \\
%     (d)&\ding{183}\ding{185}& 3.51 & 6.85 & 10.2\%  \\
%     (e)&\ding{183}\ding{185}\ding{186} &\textbf{3.36} & \textbf{6.63} &  \textbf{9.7\%} \\
%     \bottomrule[1pt]
%     \end{tabular}
%     \label{ablation}
%     \caption{Five settings on METR-LA-1 hour forecasting.}
%     \end{subfigure}
%     % \hspace{0.05\textwidth}
%     \begin{subfigure}{0.55\textwidth}
%         \centering
%         \includegraphics[width=0.7\textwidth]{figs/testing metr.png}
%         \caption{T-test on three core components.}
%     \end{subfigure}

%     \caption{Effectiveness Study of the core components of AGC-net.}
% \end{figure*}

%%%%%%%%------------ ablation
\begin{table}[htp]
\centering
\small
\caption{Ablation Study on METR-LA (1 hour forecasting).}

\begin{tabular}{c|c|ccc}
\toprule[1pt]
Setting \# & Modules &MAE & RMSE & MAPE  \\
\hline
(a)& \ding{182} &3.70 & 7.14 & 10.7\%  \\
(b)&\ding{182}\ding{183}  & 3.53 &  6.87 &  10.3\% \\
(c)&\ding{183}\ding{184} &3.58 & 6.89 & 10.5\% \\
(d)&\ding{183}\ding{185}& 3.51 & 6.85 & 10.2\%  \\
(e)&\ding{183}\ding{185}\ding{186} &\textbf{3.36} & \textbf{6.63} &  \textbf{9.7\%} \\

\bottomrule[1pt]
\end{tabular}

\centering
\label{ablation}
\end{table}

\subsection{Ablation Study}\label{effective}

We conduct ablation studies in terms of 5 different settings, i.e., (a). Using adjacency matrix (denoted as \ding{182}) for single-range graph convolution kernel. (b). Using periodic patterns (\ding{183}) introduced in \cite{guo2019attention}. (c). Using MGC-Weighted (\ding{184}) to replace convolution in setting (a). (d). Using MGC-Attention (\ding{185}) instead of MGC-Weighted. (e). Using shifted convolution kernel (\ding{186}) in the MGC-Attention.

Table \ref{ablation} demonstrates the performance under the above settings, which verify the effectiveness of each module. We analyze each observation as followings. \textbf{First}, observations from (a) verify the effectiveness of encoder-decoder architecture that obtains similar performance as \cite{li2017diffusion}. \textbf{Second}, involving periodic patterns improves performance which verifies the existence of various temporal patterns. \textbf{Third}, comparing settings (c) and (d), we observe that by using MGC-Attention, the performance increases significantly compared with MGC-Weighted. This indicates that the context attention mechanism can learn a better composition of convolutions by exploiting complex contextual information. 

At the last setting (e), the result shows that the shifted convolution kernel has a significant performance improvement. It verifies that accurate structural information is very important to the prediction task. Moreover, the shifted convolution kernel performs well even if it is composed of two low dimension matrices instead of a full-dimension matrix. This is because that under the condition that the topology is sparse in these two datasets, only a few complementary edge completions are sufficient for gaining more precise structural information. Due to the space limitation, more experimental results and discussions on hyperparameters can be found in the code repository.

\begin{figure}[t]
\centering
\includegraphics[width=0.6\textwidth]{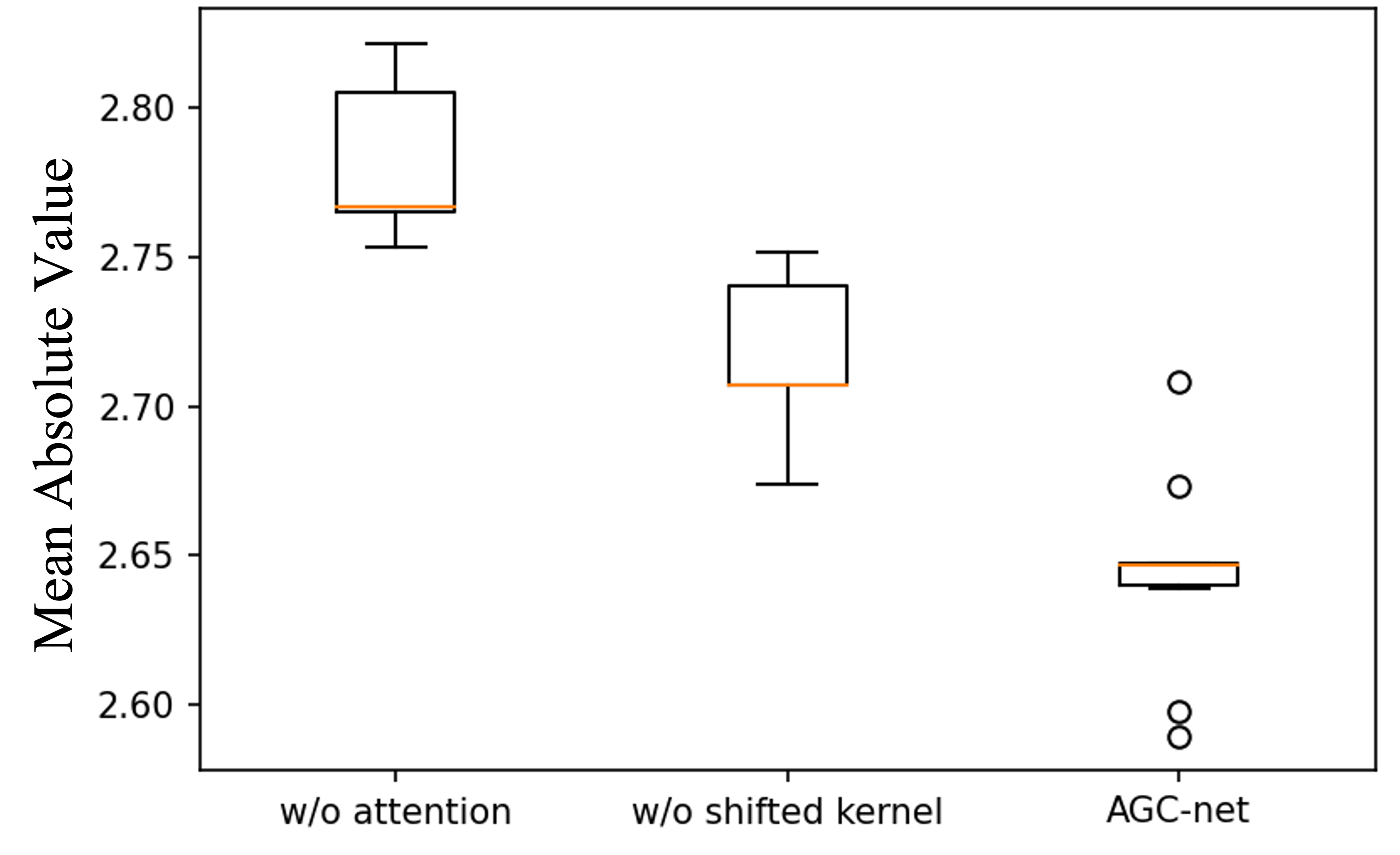}
\caption{\small T-test based significant analysis on METR-LA with 15min forecasting.}
\label{att}
\end{figure}

To further justify the effectiveness of the core component, we conducted a significance analysis using a t-test method on the three proposed modules of AGC-net, 10 rounds of training  with different random seeds were conducted on the METR-15 dataset. The results are shown in Figure 3. We found that each module significantly improves performance with a very low p-value (less than 0.001), and the small standard deviation of MAE from each setting indicates the stability of each module.

Three modules are under following different settings correspondingly:
(a) 'w/o attention' (only one fixed-range graph wavelet transform is applied, without attention and shifted kernel),
(b) 'w/o shifted kernel' (attention is applied without the shifted kernel), and
(c) 'AGC-net' (all modules are applied).

%% file: conc.tex
\section{Conclusions}
Main conclusions are as follows, (1). the experimental results verified the existence of the dynamic spatial correlations at different time steps, which is a challenge for graph convolution. The proposed adaptive graph convolution is well to extract such spatial dynamics. (2). A learnable shifted graph convolution kernel is proposed to enhance the graph convolution to obtain more accurate spatial information, and the experiments validate its effectiveness for traffic flow prediction. Moreover, the experiments show that the model has competitive performance even when the kernel is in a low dimension. (3). Experimental results demonstrate that our model outperforms baseline methods significantly.

%% file: ref.bbl
\begin{thebibliography}{10}
\providecommand{\url}[1]{\texttt{#1}}
\providecommand{\urlprefix}{URL }
\providecommand{\doi}[1]{https://doi.org/#1}

\bibitem{bai2020adaptive}
BAI, L., Yao, L., Li, C., Wang, X., Wang, C.: Adaptive graph convolutional
  recurrent network for traffic forecasting. Advances in Neural Information
  Processing Systems  \textbf{33} (2020)

\bibitem{bronstein2017geometric}
Bronstein, M.M., Bruna, J., LeCun, Y., Szlam, A., Vandergheynst, P.: Geometric
  deep learning: going beyond euclidean data. IEEE Signal Processing Magazine
  \textbf{34}(4),  18--42 (2017)

\bibitem{bruna2013spectral}
Bruna, J., Zaremba, W., Szlam, A., LeCun, Y.: Spectral networks and locally
  connected networks on graphs. arXiv preprint arXiv:1312.6203  (2013)

\bibitem{Chen2020MRA}
Chen, W., Chen, L., Xie, Y., Cao, W., Gao, Y., Feng, X.: Multi-range attentive
  bicomponent graph convolutional network for traffic forecasting. In:
  Proceedings of the AAAI Conference on Artificial Intelligence. vol.~34, pp.
  3529--3536 (2020)

\bibitem{chien2003dynamic}
Chien, S.I.J., Kuchipudi, C.M.: Dynamic travel time prediction with real-time
  and historic data. Journal of transportation engineering  \textbf{129}(6),
  608--616 (2003)

\bibitem{chung2014empirical}
Chung, J., Gulcehre, C., Cho, K., Bengio, Y.: Empirical evaluation of gated
  recurrent neural networks on sequence modeling. arXiv preprint
  arXiv:1412.3555  (2014)

\bibitem{defferrard2016convolutional}
Defferrard, M., Bresson, X., Vandergheynst, P.: Convolutional neural networks
  on graphs with fast localized spectral filtering. In: Advances in neural
  information processing systems. pp. 3844--3852 (2016)

\bibitem{deng2016latent}
Deng, D., Shahabi, C., Demiryurek, U., Zhu, L., Yu, R., Liu, Y.: Latent space
  model for road networks to predict time-varying traffic. In: Proceedings of
  the 22nd ACM SIGKDD International Conference on Knowledge Discovery and Data
  Mining. pp. 1525--1534 (2016)

\bibitem{diao2019dynamic}
Diao, Z., Wang, X., Zhang, D., Liu, Y., Xie, K., He, S.: Dynamic
  spatial-temporal graph convolutional neural networks for traffic forecasting.
  In: Proceedings of the AAAI conference on artificial intelligence. vol.~33,
  pp. 890--897 (2019)

\bibitem{gao2018large}
Gao, H., Wang, Z., Ji, S.: Large-scale learnable graph convolutional networks.
  In: Proceedings of the 24th ACM SIGKDD International Conference on Knowledge
  Discovery \& Data Mining. pp. 1416--1424 (2018)

\bibitem{guo2021hierarchical}
Guo, K., Hu, Y., Sun, Y., Qian, S., Gao, J., Yin, B.: Hierarchical graph
  convolution network for traffic forecasting. In: Proceedings of the AAAI
  Conference on Artificial Intelligence. vol.~35, pp. 151--159 (2021)

\bibitem{guo2019attention}
Guo, S., Lin, Y., Feng, N., Song, C., Wan, H.: Attention based spatial-temporal
  graph convolutional networks for traffic flow forecasting. In: Proceedings of
  the AAAI Conference on Artificial Intelligence. vol.~33, pp. 922--929 (2019)

\bibitem{hamilton2017inductive}
Hamilton, W., Ying, Z., Leskovec, J.: Inductive representation learning on
  large graphs. In: Advances in neural information processing systems. pp.
  1024--1034 (2017)

\bibitem{hammond2011wavelets}
Hammond, D.K., Vandergheynst, P., Gribonval, R.: Wavelets on graphs via
  spectral graph theory. Applied and Computational Harmonic Analysis
  \textbf{30}(2),  129--150 (2011)

\bibitem{kingma2014adam}
Kingma, D.P., Ba, J.: Adam: A method for stochastic optimization. arXiv
  preprint arXiv:1412.6980  (2014)

\bibitem{kipf2016semi}
Kipf, T.N., Welling, M.: Semi-supervised classification with graph
  convolutional networks. arXiv preprint arXiv:1609.02907  (2016)

\bibitem{li2021spatial}
Li, M., Zhu, Z.: Spatial-temporal fusion graph neural networks for traffic flow
  forecasting. In: Proceedings of the AAAI Conference on Artificial
  Intelligence. vol.~35, pp. 4189--4196 (2021)

\bibitem{li2017diffusion}
Li, Y., Yu, R., Shahabi, C., Liu, Y.: Diffusion convolutional recurrent neural
  network: Data-driven traffic forecasting. arXiv preprint arXiv:1707.01926
  (2017)

\bibitem{liu2016predicting}
Liu, Q., Wu, S., Wang, L., Tan, T.: Predicting the next location: A recurrent
  model with spatial and temporal contexts. In: Proceedings of the AAAI
  Conference on Artificial Intelligence. vol.~30 (2016)

\bibitem{makridakis1997arma}
Makridakis, S., Hibon, M.: Arma models and the box--jenkins methodology.
  Journal of Forecasting  \textbf{16}(3),  147--163 (1997)

\bibitem{niepert2016learning}
Niepert, M., Ahmed, M., Kutzkov, K.: Learning convolutional neural networks for
  graphs. In: International conference on machine learning. pp. 2014--2023
  (2016)

\bibitem{oreshkin2021fc}
Oreshkin, B.N., Amini, A., Coyle, L., Coates, M.: Fc-gaga: Fully connected
  gated graph architecture for spatio-temporal traffic forecasting. In:
  Proceedings of the AAAI Conference on Artificial Intelligence. vol.~35, pp.
  9233--9241 (2021)

\bibitem{shekhar2015spatiotemporal}
Shekhar, S., Jiang, Z., Ali, R.Y., Eftelioglu, E., Tang, X., Gunturi, V., Zhou,
  X.: Spatiotemporal data mining: A computational perspective. ISPRS
  International Journal of Geo-Information  \textbf{4}(4),  2306--2338 (2015)

\bibitem{stathopoulos2003multivariate}
Stathopoulos, A., Karlaftis, M.G.: A multivariate state space approach for
  urban traffic flow modeling and prediction. Transportation Research Part C:
  Emerging Technologies  \textbf{11}(2),  121--135 (2003)

\bibitem{sun2022ada}
Sun, J., Li, J., Wu, C., Tang, Z., Wu, C.: Ada-stnet: A dynamic adaboost
  spatio-temporal network for traffic flow prediction. In: ICASSP 2022-2022
  IEEE International Conference on Acoustics, Speech and Signal Processing
  (ICASSP). pp. 5478--5482. IEEE (2022)

\bibitem{sun2006bayesian}
Sun, S., Zhang, C., Yu, G.: A bayesian network approach to traffic flow
  forecasting. IEEE Transactions on intelligent transportation systems
  \textbf{7}(1),  124--132 (2006)

\bibitem{tian2021spatial}
Tian, C., Chan, W.K.: Spatial-temporal attention wavenet: A deep learning
  framework for traffic prediction considering spatial-temporal dependencies.
  IET Intelligent Transport Systems  \textbf{15},  549--561 (2021)

\bibitem{wu2004travel}
Wu, C.H., Ho, J.M., Lee, D.T.: Travel-time prediction with support vector
  regression. IEEE transactions on intelligent transportation systems
  \textbf{5}(4),  276--281 (2004)

\bibitem{wu2016urban}
Wu, Y.J., Chen, F., Lu, C.T., Yang, S.: Urban traffic flow prediction using a
  spatio-temporal random effects model. Journal of Intelligent Transportation
  Systems  \textbf{20},  282--293 (2016)

\bibitem{Wu2019GWN}
Wu, Z., Pan, S., Long, G., Jiang, J., Zhang, C.: Graph wavenet for deep
  spatial-temporal graph modeling. In: Proceedings of the Twenty-Eighth
  International Joint Conference on Artificial Intelligence, {IJCAI-19}. pp.
  1907--1913 (2019)

\bibitem{xu2019graph}
Xu, B., Shen, H., Cao, Q., Qiu, Y., Cheng, X.: Graph wavelet neural network.
  arXiv preprint arXiv:1904.07785  (2019)

\bibitem{xu2020spatial}
Xu, M., Dai, W., Liu, C., Gao, X., Lin, W., Qi, G.J., Xiong, H.:
  Spatial-temporal transformer networks for traffic flow forecasting. arXiv
  preprint arXiv:2001.02908  (2020)

\bibitem{zhang2017deep}
Zhang, J., Zheng, Y., Qi, D.: Deep spatio-temporal residual networks for
  citywide crowd flows prediction. In: Proceedings of the AAAI Conference on
  Artificial Intelligence. vol.~31 (2017)

\bibitem{zhang2020spatio}
Zhang, Q., Chang, J., Meng, G., Xiang, S., Pan, C.: Spatio-temporal graph
  structure learning for traffic forecasting. In: Proceedings of the AAAI
  Conference on Artificial Intelligence. vol.~34, pp. 1177--1185 (2020)

\bibitem{zhang2021traffic}
Zhang, X., Huang, C., Xu, Y., Xia, L., Dai, P., Bo, L., Zhang, J., Zheng, Y.:
  Traffic flow forecasting with spatial-temporal graph diffusion network. In:
  Proceedings of the AAAI Conference on Artificial Intelligence. vol.~35, pp.
  15008--15015 (2021)

\bibitem{zheng2020gman}
Zheng, C., Fan, X., Wang, C., Qi, J.: Gman: A graph multi-attention network for
  traffic prediction. In: Proceedings of the AAAI Conference on Artificial
  Intelligence. vol.~34, pp. 1234--1241 (2020)

\bibitem{zhou2020graph}
Zhou, J., Cui, G., Hu, S., Zhang, Z., Yang, C., Liu, Z., Wang, L., Li, C., Sun,
  M.: Graph neural networks: A review of methods and applications. AI Open
  \textbf{1},  57--81 (2020)

\bibitem{zivot2006vector}
Zivot, E., Wang, J.: Vector autoregressive models for multivariate time series.
  Modeling financial time series with S-PLUS{\textregistered} pp. 385--429
  (2006)

\end{thebibliography}
